\def\year{2022}\relax
\documentclass[letterpaper]{article} 
\usepackage{aaai22}  
\usepackage{times}  
\usepackage{helvet}  
\usepackage{courier}  
\usepackage[hyphens]{url}  
\usepackage{graphicx} 
\usepackage{natbib}  
\usepackage{caption} 
\usepackage{hyperref}
\DeclareCaptionStyle{ruled}{labelfont=normalfont,labelsep=colon,strut=off} 
\usepackage{algorithm}
\usepackage{algorithmic}

\usepackage{mathtools}              
\usepackage{enumitem}     
\usepackage{subcaption}

\newcommand{\taskname}{{\sc MuMuQA}}

\usepackage{xparse}
\NewDocumentCommand{\heng}
{ mO{} }{\textcolor{red}{\textsuperscript{\textit{Heng}}\textsf{\textbf{\small[#1]}}}}
\NewDocumentCommand{\shifu}
{ mO{} }{\textcolor{Cyan}{\textsuperscript{\textit{Shi-Fu}}\textsf{\textbf{\small[#1]}}}}
\NewDocumentCommand{\revanth}
{ mO{} }{\textcolor{brown}{\textsuperscript{\textit{Revanth}}\textsf{\textbf{\small[#1]}}}}
\NewDocumentCommand{\manling}
{ mO{} }{\textcolor{CadetBlue}{\textsuperscript{\textit{Manling}}\textsf{\textbf{\small[#1]}}}}
\NewDocumentCommand{\haoyang}
{ mO{} }{\textcolor{Turquoise}{\textsuperscript{\textit{Haoyang}}\textsf{\textbf{\small[#1]}}}}
\NewDocumentCommand{\xudong}
{ mO{} }{\textcolor{Green}{\textsuperscript{\textit{Xudong}}\textsf{\textbf{\small[#1]}}}}
\NewDocumentCommand{\ali}
{ mO{} }{\textcolor{Serenity}{\textsuperscript{\textit{Ali}}\textsf{\textbf{\small[#1]}}}}
\NewDocumentCommand{\avi}{ mO{} }{\textcolor{orange}{\textsuperscript{\textit{Avi}}\textsf{\textbf{\small[#1]}}}}
\NewDocumentCommand{\lifu}{ mO{} }{\textcolor{blue}{\textsuperscript{\textit{Lifu}}\textsf{\textbf{\small[#1]}}}}
\NewDocumentCommand{\jc}{ mO{} }{\textcolor{violet}{\textsuperscript{\textit{Jaemin}}\textsf{\textbf{\small[#1]}}}}
\NewDocumentCommand{\mbc}{ mO{} }{\textcolor{red}{\textsuperscript{\textit{Mohit}}\textsf{\textbf{\small[#1]}}}}
\NewDocumentCommand{\as}{ mO{} }{\textcolor{magenta}{\textsuperscript{\textit{Alex}}\textsf{\textbf{\small[#1]}}}}

\usepackage{xcolor}
\definecolor{ceruleanblue}{rgb}{0.16, 0.32, 0.75}

\usepackage{newfloat}
\usepackage{listings}
\floatstyle{ruled}
\newfloat{listing}{tb}{lst}{}
\floatname{listing}{Listing}
\title{MuMuQA: Multimedia Multi-Hop News Question Answering via
\\ Cross-Media Knowledge Extraction and Grounding}
\author{
   Revanth Gangi Reddy\textsuperscript{\rm 1}, Xilin Rui\textsuperscript{\rm 2}, Manling Li\textsuperscript{\rm 1}, Xudong Lin\textsuperscript{\rm 3}, Haoyang Wen\textsuperscript{\rm 1}, Jaemin Cho\textsuperscript{\rm 4},\\ 
   Lifu Huang\textsuperscript{\rm 5}, Mohit Bansal\textsuperscript{\rm 4}, Avirup Sil\textsuperscript{\rm 6},
   Shih-Fu Chang\textsuperscript{\rm 3}, Alexander Schwing\textsuperscript{\rm 1}, Heng Ji\textsuperscript{\rm 1}
}
\affiliations{
    \textsuperscript{\rm 1} University of Illinois Urbana-Champaign \hspace{1em} \textsuperscript{\rm 2} Tsinghua University\hspace{1em} \textsuperscript{\rm 3} Columbia University \\ 
    \textsuperscript{\rm 4} University of North Carolina at Chapel Hill \hspace{1em}
    \textsuperscript{\rm 5} Virginia Tech \hspace{1em} \textsuperscript{\rm 6} IBM Research AI \\
    \texttt{\{revanth3,hengji\}@illinois.edu}
}

\begin{document}

\maketitle

\begin{abstract}

Recently, there has been an increasing interest in building question answering (QA) models that reason across multiple modalities, such as text and images. However, QA using images is often limited to just picking the answer from a pre-defined set of options.
In addition, images in the real world, especially in news, have objects that are co-referential to the text, with complementary information from both modalities. 
In this paper, we present a new QA evaluation benchmark with 1,384 questions over news articles that require \textit{cross-media grounding} of objects in images onto text.
Specifically, the task involves multi-hop questions that require reasoning over image-caption pairs to identify the grounded visual object being referred to and then predicting a span from the news body text to answer the question. In addition, we introduce a novel multimedia data augmentation framework, based on cross-media knowledge extraction and synthetic question-answer generation, to automatically augment data that can provide weak supervision for this task. We evaluate both pipeline-based and end-to-end pretraining-based multimedia QA models on our benchmark, and show that they achieve promising performance, while considerably lagging behind human performance hence leaving large room for future work on this challenging new task.\footnote{All the datasets, programs and tools will be made publicly available here: \href{https://github.com/uiucnlp/MuMuQA}{https://github.com/uiucnlp/MuMuQA}}

\end{abstract}

\section{Introduction}

\begin{figure*}[ht]
     \begin{subfigure}[b]{0.48\textwidth}       
     \includegraphics[scale=0.36]{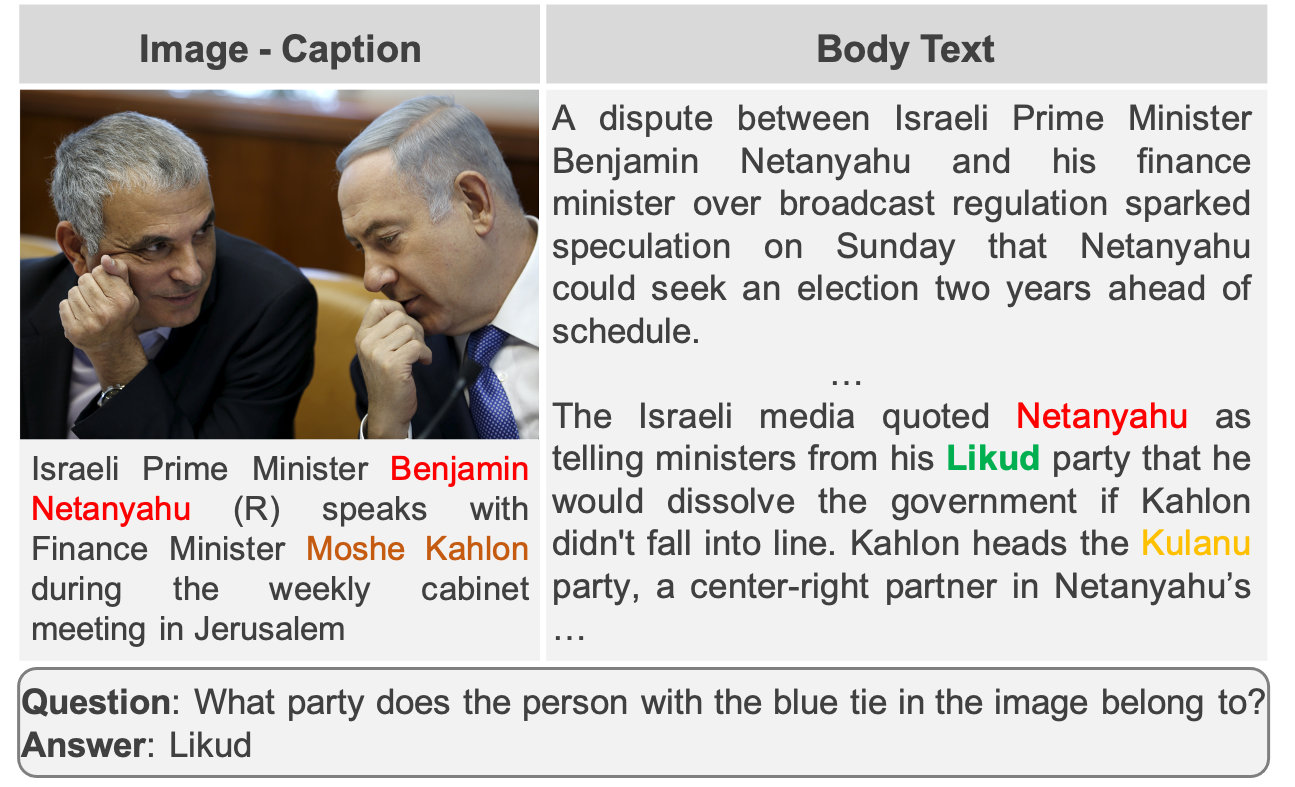}       \caption{}
       \label{fig:mmqa_example_1}
     \end{subfigure}
     \hfill
     \begin{subfigure}[b]{0.48\textwidth}       
     \includegraphics[scale=0.22]{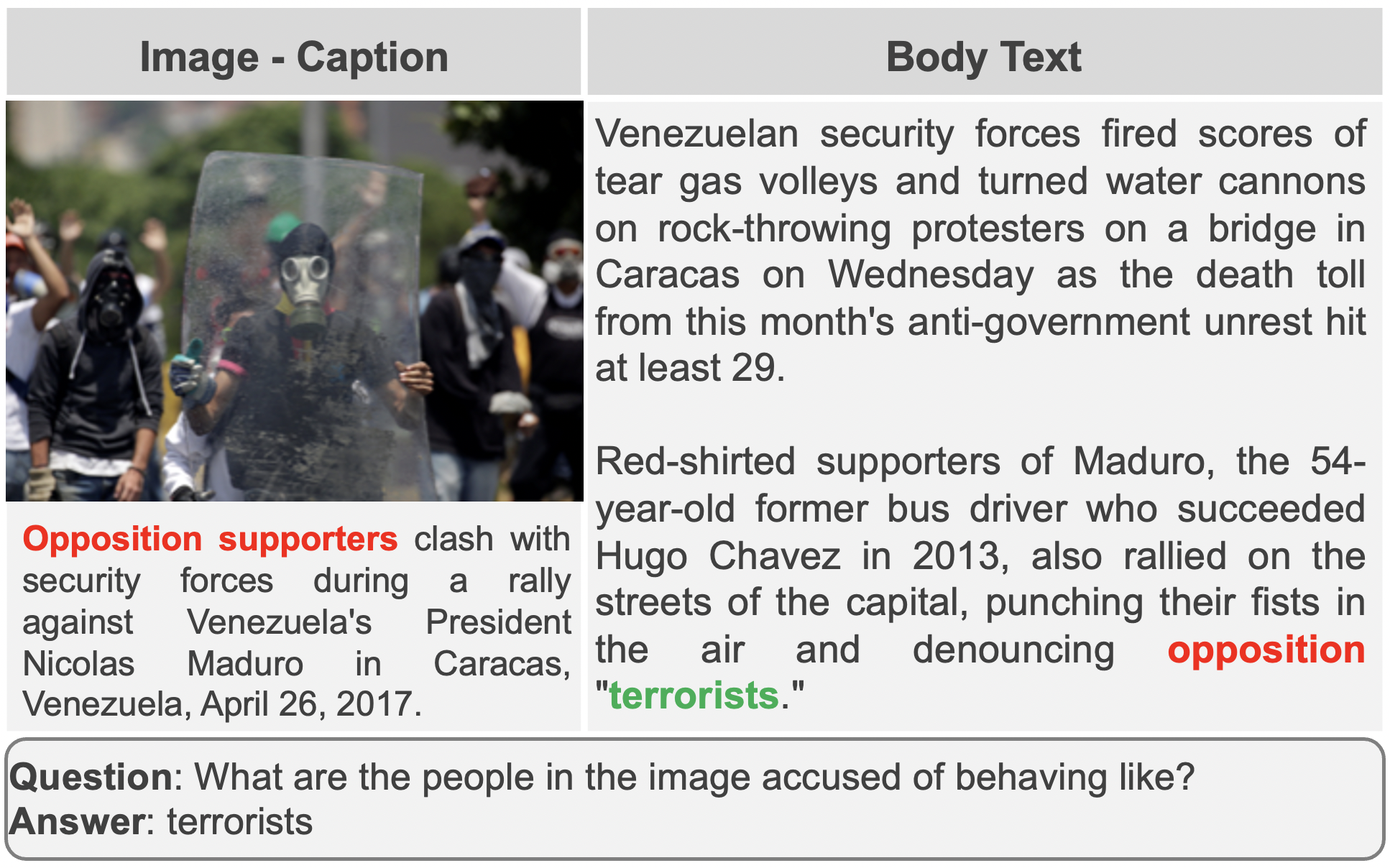}       \caption{}
       \label{fig:mmqa_example_2}
     \end{subfigure}
     \caption{Two examples from our evaluation benchmark with the question-answer pairs and their corresponding news articles. The bridge item, which needs to be grounded from image onto text, is shown in red and the final answer is marked in green in the news body text.
     }
     \label{fig:walkthrough_example}
\end{figure*}

To answer questions, humans seamlessly combine context provided in the form of images, text, background knowledge or structured data such as graphs and tables. Availability of a similarly capable question answering (QA) model could increase information accessibility and allow 
quick understanding of news or scientific articles, where the main narrative is enriched by images, charts and graphs with captions.

However, most QA research to date focuses on extracting information from a single data modality, such as text (TextQA) \cite{rajpurkar2016squad, kwiatkowski2019natural}, images (VQA) \cite{antol2015vqa, goyal2017making} or videos (VideoQA) \cite{yang2003videoqa}. Only recently, there have been initial attempts to combine information from multiple modalities \cite{kembhavi2017you,lei2018tvqa}, which often use questions in a multiple-choice setting. However, multiple choice questions in VQA have been shown to have explicit bias \cite{kazemi2017show, manjunatha2019explicit}, which can be taken advantage of by using question understanding or answer option analysis.

To address this concern, more recently, \citet{talmor2021multimodalqa} introduce an extractive multi-hop QA dataset that involves reasoning across tables, text and images. However, each image is associated with a Wikipedia entity, therefore the reasoning over images essentially boils down to ranking the images based on the question and using the entities corresponding to the top-ranked images. 
In contrast, questions about images in real life require cross-media reasoning between images and text. Current multimedia benchmarks cannot be used to solve this as they do not require to ground images onto text when answering questions about images.

To bridge this gap, we introduce a new benchmark QA evaluation task called `Multimedia Multi-hop Question Answering' (\taskname{}) along with a corresponding dataset of news articles. To succeed in this new \taskname{} task, methods need to excel at both identifying objects being referred to in the question and leveraging news body text to answer the question. Specifically, the task is formulated as multi-hop extractive QA where
questions focus on objects grounded in  image-caption pairs. To enforce multi-hop reasoning, we ask information-seeking questions about objects in the image by referring to them via their visual attributes. The requirement for grounding between image and text along with multi-hop reasoning makes \taskname{} much more challenging than the tasks which are commonly explored in the QA community so far.

Figure~\ref{fig:walkthrough_example} illustrates the task and the data. Answering the question in Figure~\ref{fig:mmqa_example_1} requires first identifying that ``the person with the blue tie'' is on the right in the image, which needs to be grounded to ``Benjamin Netanyahu'' (in red) in the caption. Subsequently, the news body text is needed to  extract the answer. Grounding between image-caption pairs along with textual understanding and coreference resolution is hence crucial. Note that the question cannot be  answered by a text-only model, since the text mentions two parties  (green and yellow highlight) that correspond to the two people in the image-caption pair (names in red and brown). Figure~\ref{fig:mmqa_example_2} is another example, with the image necessary to disambiguate that ``people in the image" refers to \textit{opposition supporters} and not \textit{security forces}.

To study \taskname{}, we release an evaluation set with 263 development and 1,121 test examples. Annotators were asked to identify objects grounded in image-caption pairs and come up with questions about them which can be answered by the news body text. Given the high cost of annotating such examples, we use human-curated examples only as the evaluation set and develop a novel multimedia data augmentation approach to automatically generate silver-standard training data for this task, which can then be used to train or fine-tune state-of-the-art (SOTA) vision-language models \cite{tan2019lxmert, li2020oscar}. Specifically, we generate silver training data by leveraging multimedia knowledge extraction, visual scene understanding and language generation. In short, we first run a state-of-the-art multimedia knowledge extraction system \cite{li2020gaia} to capture the entities that are grounded in image-caption pairs, such as ``Benjamin Netanyahu''. Next, we apply a question generation approach \cite{shakeri2020end} to automatically generate questions from the news body text, that are conditioned on one of the grounded entities, such as ``What party does Benjamin Netanyahu belong to?''. Then, we use a visual attribute recognition model \cite{lin2019improving} to edit these questions to refer to the grounded entity by its visual attributes, such as replacing ``Benjamin Netanyahu'' with ``the person with the blue tie'' in the question. Finally, we filter out questions that are answerable by a single-hop text-only QA model \cite{chakravarti2020towards}.
This pipeline automatically generates 
training data for our task.

We evaluate both text-only QA and multimedia QA models on our \taskname{} benchmark. Particularly, we explore both pipeline-based and end-to-end multimedia QA approaches. The pipeline-based model first decomposes the task into image-question and text-question. Then, it uses a SOTA multimedia knowledge extraction model~\cite{li2020gaia} for visual entity grounding and attribute matching to answer the image-question and uses a single-hop text QA model~\cite{chakravarti2020towards} to extract the final answer. For the end-to-end multimedia QA system, we directly finetune a SOTA visual-language pretrained model~\cite{li2020oscar} using auto-generated silver-standard training data.

The contributions of this work are as follows:
\begin{itemize}
    \item We release a new extractive QA evaluation benchmark, \taskname{}. It is based on multi-hop reasoning and cross-media grounding of information present in news articles. 
   To the best of our knowledge, our work is the first to attempt using information grounded in the image in an extractive QA setting (see Section~\ref{sec:task} and Section~\ref{sec:data}).
   
    \item To automatically generate silver-standard training data for this task, we introduce a novel pipeline that incorporates cross-media knowledge extraction, visual understanding and synthetic question generation (Section~\ref{sec:syn_data}). 
    
    \item We measure the impact of images in our benchmark by evaluating competitive text-only QA models on our task, and
    demonstrate the benefit of using multimedia information (see Section~\ref{sec:QAModels}).
    
\end{itemize}

\section{\taskname{} Task}
\label{sec:task}

In this section, we present the details of our Multimedia Multi-hop Question Answering (\taskname {}) task. As illustrated in Figure~\ref{fig:walkthrough_example}, given a news article with an image-caption pair and a question, a system needs to answer the question by extracting a short span from the body text.
Importantly, answering the questions requires multi-hop reasoning: the first hop, referred to as \textit{image entity grounding}, requires cross-media grounding between the image and caption to obtain an intermediate answer, named \textit{bridge item}, for the image-related question; and the second hop requires reasoning over the news body text by using the bridge item to extract a span of text 
as the final answer. For example, in Figure~\ref{fig:mmqa_example_1}, with the first hop, we need to ground ``person with the blue tie in the image'' to the particular entity ``Benjamin Netanyahu'' in the caption. Taking ``Benjamin Netanyahu'' as the bridge item to the second hop, we further extract the final answer as ``Likud'' from the news body text. The questions require using information present in the image for entity disambiguation, thereby needing cross-media grounding. 

Our benchmark reflects questions that news readers might have after looking at the visual information in the news article, without having read the relatively longer body text. News articles usually have objects that are mentioned in both images and text, thereby requiring cross-media grounding to answer questions about them. In our task, we follow the ordering of using the visual information first, and then pick answer from the news body text so as to allow for a wide range of answers. In contrast, following the other reasoning order would require answers to come from images, which previous VQA work has shown to be restricted to a pre-defined vocabulary. We use multi-hop questions to enforce the constraint of using information from different modalities, which also follows recent work on multimedia QA~\cite{talmor2021multimodalqa}, that however does not require any cross-media grounding. Prior work~\cite{yang2018hotpotqa, welbl2018constructing} has also used multi-hop questions as a means to evaluate such complex reasoning chains.

\section{Benchmark Construction}

\label{sec:data}

To facilitate progress on this multi-hop extractive \taskname{} task, we collect a new dataset. Our dataset consists of an evaluation set that is human-annotated and a silver-standard training set that is automatically generated (described later in Section~\ref{sec:syn_data}). 
We choose to manually construct an evaluation set of high quality, to mimick the information-seeking process of real newsreaders. 
We first provide a brief description of the news articles in our dataset (Section~\ref{sec:data:articles}) and then detail how the evaluation set (Section~\ref{sec:data:eval}) was created, along with a brief analysis (Section~\ref{sec:analysis}).

\subsection{News Data Acquisition}
\label{sec:data:articles}

We take 108,693 news articles (2009-2017) from the Voice of America (VOA) website\footnote{\url{www.voanews.com}}, covering a wide array of topics such as military, economy and health. We use articles from 2017 for annotating the evaluation set and articles from 2009 to 2016 for creating the training set. 

\subsection{Evaluation Set Construction}
\label{sec:data:eval}

To evaluate models, we collect data via a newly developed and specifically tailored annotation interface\footnote{Screenshot of the interface and annotation instructions are available in the appendix. The annotation tool will be made publicly available.}. News articles are shown in the interface along with their images and corresponding captions and annotators are asked to come up with questions. The  annotation process requires the annotator to first look at the image-caption pair to identify which objects in the image are grounded in the caption and to choose one of them as the bridge item. The annotators then study the news body text to look for mentions of the bridge item and pick a sentence for which the question will be created. Finally, annotators create a multi-hop question with the answer coming from the news body text and the bridge item being referred to in the question by its visual attributes. 

To ensure a real need for multimedia reasoning, we provide annotators with access to a publicly available single-hop text-only QA model \cite{chakravarti2020towards}. Specifically, annotators are asked to ensure that the text-only model cannot pick up the answer by looking at only the question and the news body text. 
We also ensure that the annotators do not attempt to `beat' the text-only model by formulating questions that are too complex or convoluted. For this, we require the answer to appear in the top-5 text-only model answers when the image reference is removed from the question, by replacing it with the \textit{bridge item}. 
Additionally, note that the annotators are specifically asked to not use their own background knowledge of the people in the images when trying to ground them into the caption, i.e.,  grounding is intended to be purely based on visual or spatial attributes, without any need for explicit face recognition.  

Our annotators are native English speakers and have had extensive experience annotating for other NLP tasks. On average, they took 3.5 
minutes to come up with the question and annotated one in every three news articles they saw (i.e., they chose to skip the other two). 

For quality control, we  ask a different set of senior annotators to validate all examples in the evaluation set. Specifically, we ask to check that the image reference is grounded in the caption, questions are answerable and unambiguous. \citet{kwiatkowski2019natural} have shown that aggregating answers from multiple annotators is more robust than relying on a single annotator. Following them, we use an Amazon Mechanical Turk~\cite{buhrmester2016amazon} task where crowd-workers were asked to answer the question and also provide the bridge item, given the news article and the corresponding image-caption pair. We obtain one more answer per question this way, which we combine with the original answer to get 2-way annotations for each example in the evaluation set. The Turkers on average needed 2.5 minutes to answer a question. 

\subsection{Dataset Analysis}
\label{sec:analysis}
\begin{figure}[!htb]
    \centering
    \includegraphics[scale=0.13]{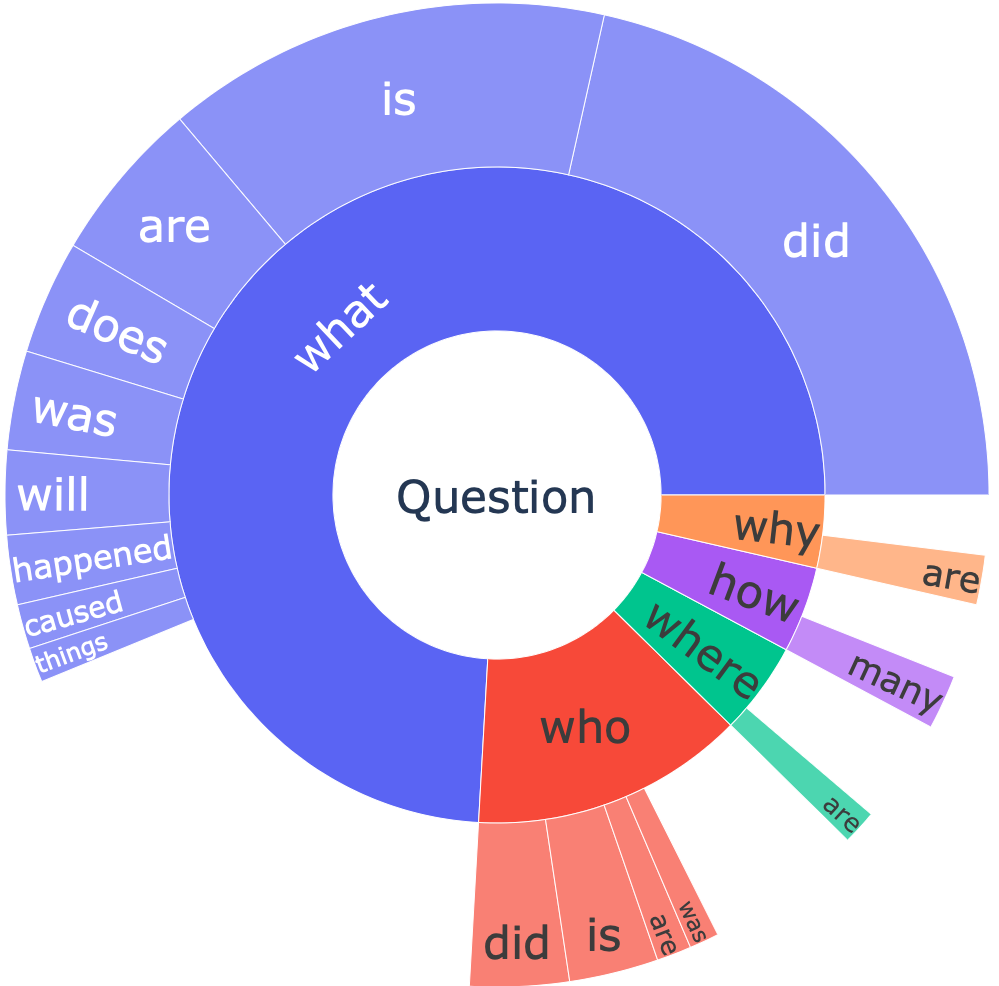}
    \caption{Distribution of questions for 5 most common first words and the subsequent second words.} 
    \label{fig:question_dist}
\end{figure}
The evaluation benchmark contains 1384 human-annotated instances, with 263 instances in the development set and the remaining in the test set. Figure~\ref{fig:question_dist} shows the sunburst plots from the analysis of the most common words in each question. We see that ``what" is the most common question type, which is similar to 
other QA datasets~\cite{lewis2020mlqa, hannan2020manymodalqa}.

\begin{figure*}[ht]
    \center
     \includegraphics[scale=0.87]{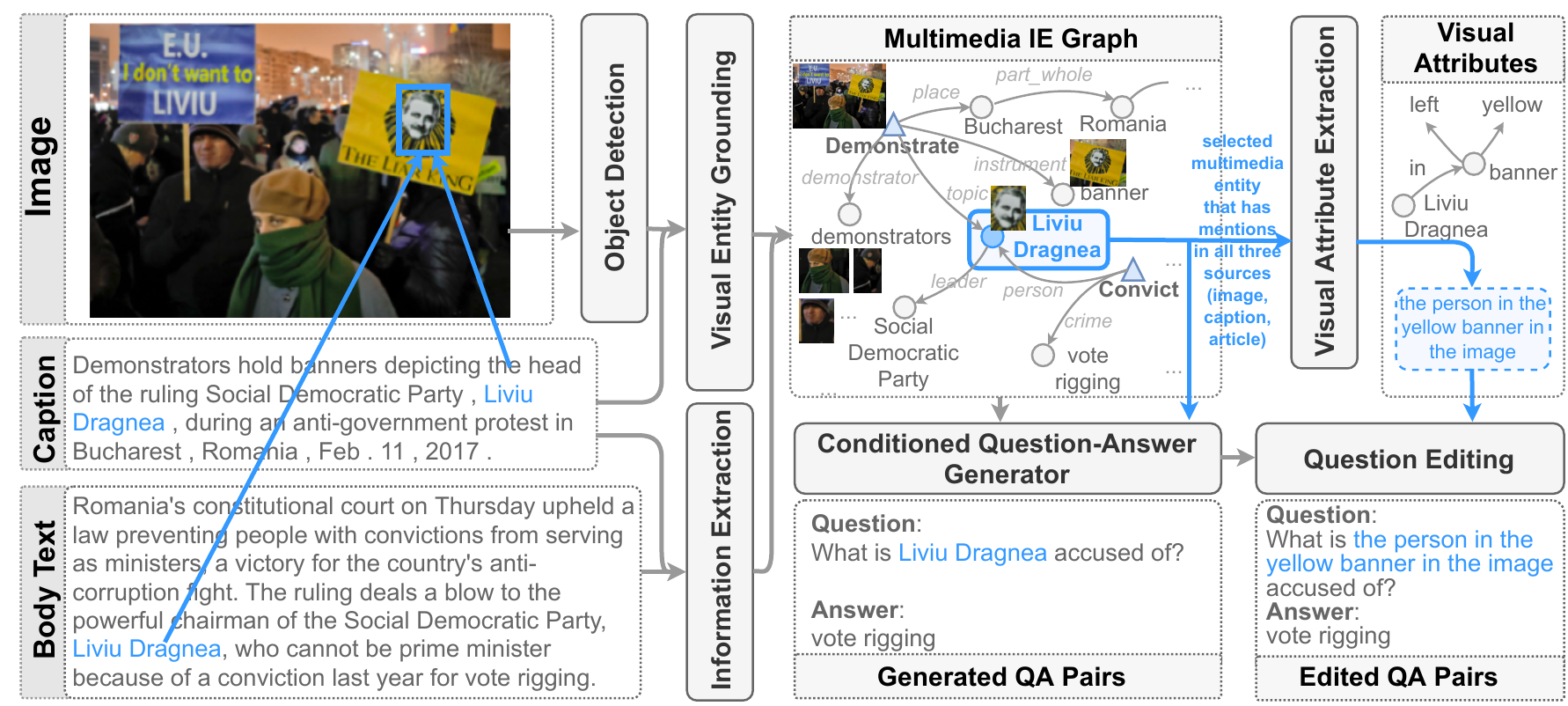}  
     \caption{The workflow of synthetic question generation for creating the training set.}
     \label{fig:qageneration}
\end{figure*}

We also analyze the entity type of the bridge item, which is the answer to the part of the multi-hop question corresponding to the image, i.e., the \textit{image-question}. We see that the majority of questions are about people (69\%), with significant questions referring to locations (10\%), organizations (10\%), nationalities or political groups (7\%) and other types (4\%). Also, we find that the bridge item can be found in the caption 89\% of the time and it is present in the news body text in the remaining cases. 

We compare the total answer vocabulary for \taskname{} with VQA and also vocabulary size per question to account for difference in dataset sizes. We see that VQA, with 214k questions and 1.5 words per answer on average, has a vocabulary size of 26k (0.12 per question). \taskname{}, with 1384 questions and 6 words per answer on average, has a vocabulary size of 3k (2.18 per question). 

\section{Silver Training Set Generation}
\label{sec:syn_data}

Given the cost and complexity associated with creating question-answer pairs required for \taskname{}, we use human-curated data only for the evaluation set. However, to support end-to-end training of models for this task, we create silver-standard training data using a novel multimedia data augmentation pipeline to generate multi-hop questions that require cross-media grounding. An overview of our generation pipeline is detailed in Section~\ref{sec:gen_process}, which consists of using multimedia entity grounding (Section~\ref{sec:grounding}), visual understanding (Section~\ref{sec:visual_attr}) and conditioned question generation (Section~\ref{sec:synth_gen}) with question editing and filtering (Section~\ref{sec:ques_edit}) to obtain questions that are multi-hop and require cross-media grounding to answer them.  

\subsection{Training Data Generation Overview}
\label{sec:gen_process}

\figurename~\ref{fig:qageneration} shows our training data generation framework. 
At a high level, we intend to automatically generate training data that contains questions about entities that are grounded in the image-caption pairs (e.g., ``Liviu Dragnea''), with the answer coming from the news body text, such as ``vote rigging''.

First, we perform multimedia entity grounding on the image-caption pairs to identify objects in the image which are grounded in the captions, to obtain the grounded entities, such as ``Liviu Dragnea'' in \figurename~\ref{fig:qageneration}. We extract the visual attributes for the grounded entities by running the visual attribute extraction system on their bounding boxes, e.g., ``yellow, banner'' in \figurename~\ref{fig:qageneration}. It enables us to generate a description ``the person in the yellow banner in the image'' for the grounded entity ``Liviu Dragnea''.

Next, we generate the questions for the grounded entities, such as ``What is Liviu Dragnea accused of?''. We first run a state-of-the-art knowledge extraction system \cite{li2020gaia} on the caption and body text to identify mentions of the grounded entities in the body text, such as ``Liviu Dragnea'' and ``chairman'' in the body text. It enables us to find candidate context for question generation, which we feed with the grounded entity $e$ into the synthetic question generator to get a question-answer pair $(q,a)$. We ensure that the generated question has a mention of the grounded entity $e$ in its text.
Then, we edit these questions to replace the grounded mention by its corresponding visual attributes, to produce the final multi-hop question such as ``What is the person in the yellow banner in the image accused of ?''.

\subsection{Multimedia Entity Grounding}
\label{sec:grounding}

We ground each entity in the text 
to a bounding box in the image using a multi-layer attention mechanism following \citet{akbari2019multi}. The system\footnote{\url{https://hub.docker.com/r/gaiaaida/grounding-merging}} extracts a multi-level visual feature map for each image in a document, where each visual feature level represents a certain semantic granularity, such as \textit{word}, \textit{phrase}, \textit{sentence}, etc. In addition, each entity is represented using contextualized embeddings~\cite{peters2018deep}, and we compute an attention map to every visual feature level and every location of the feature map. In this way, we can choose the visual feature level that most strongly matches the semantics of the entity, and the attention map can be used to localize the entity. The model is fine-tuned on the VOA news that we collected.

\subsection{Visual Attribute Extraction}
\label{sec:visual_attr}
By analyzing the object detection results, we observe that on the development set, about 75\% of images contain at least one person that can be grounded to its text mentions. Therefore, describing the attributes of a person is of great importance. 
Given the object detection results from the grounding step, we explore three types of attributes to comprehensively describe people in images: spatial attributes, personal accessories, and personal attributes.
The spatial attribute is the relative position of the person. We associate personal accessories from object detection results to the person bounding boxes. To obtain personal attributes, we use a widely-used person attribute recognition model\footnote{\url{https://github.com/hyk1996/Person-Attribute-Recognition-MarketDuke}} \cite{lin2019improving}. More details are provided in the supplementary material.

\subsection{Conditioned Question Generation}
\label{sec:synth_gen}
Given a passage in the news body text and an entity present in that passage, we aim to generate a synthetic question about that entity using the passage. Specifically, we train a synthetic example generator to take a passage $p$, an entity $e$ and generate a question $q$ and its corresponding answer $a$. To achieve this, we fine-tune BART \cite{lewis2020bart}, an encoder-decoder based model, using  Natural Questions (NQ) \cite{kwiatkowski2019natural}, an existing machine reading comprehension dataset. This dataset appears in the form of $(q,p,a)$ triples. We identify the entity $e$ in the question, that also appears in the passage. This entity is passed as input along with the passage to condition the question generation.

\begin{figure*}[ht]
    \center
     \includegraphics[scale=0.84]{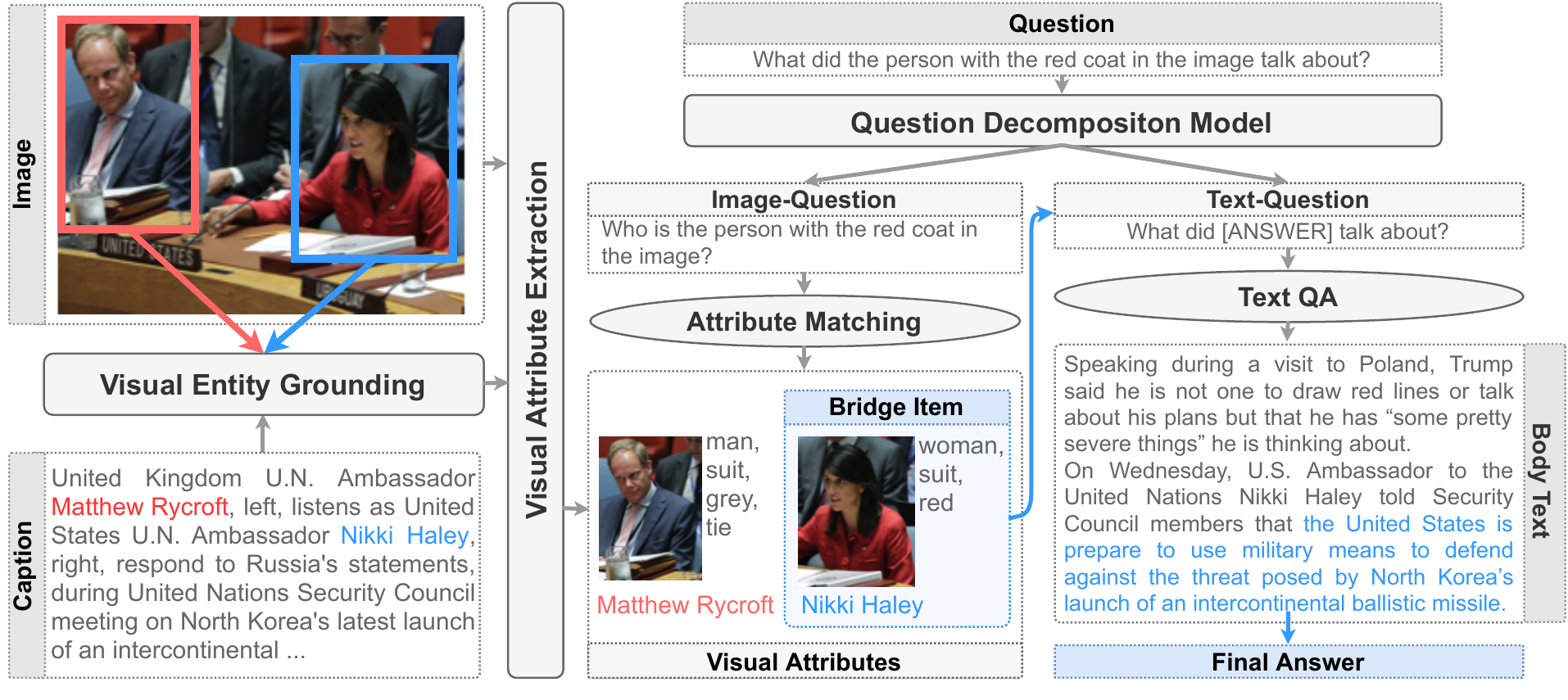}  
     \caption{Overview diagram of the pipeline based multimedia QA approach.}
     \label{fig:mmqa_pipeline}
\end{figure*}

\subsection{Question Editing and Filtering}
\label{sec:ques_edit}
We edit the questions by replacing the grounded entity's text mention in the question with their visual attributes. This step aims to incorporate the need for using the image to do entity disambiguation. Finally, to filter out questions that are answerable directly using the text, we apply a cycle-consistency filter similar to \citet{alberti2019synthetic}, by using a single-hop text-only QA model \cite{chakravarti2020towards}. This step discards questions that can be answered using text-only cues, so as to ensure that information from the image is required for answering the question. Since synthetic questions can be noisy, we also discard questions for which the answer does not appear in the top-5 answers to ensure quality of questions.

\section{Question Answering Models}
\label{sec:QAModels}

In this section, we describe various competitive baseline methods for evaluation on our benchmark. 
We begin with a multi-hop text-only extractive QA baseline and then proceed to use both pipeline-based and end-to-end multimedia {QA} baselines.

\subsection{Multi-hop Text-only QA}
\label{sec:text_only}

The multi-hop text-only QA model takes the question, caption and body text as input, and processes each paragraph in the body text independently, similar to \citet{alberti2019bert}.
The QA model has an extractive answer predictor that predicts the indices of answer spans on top of the output representations $\mathbf{H}$ from pre-trained language models (LM) \cite{devlin2019bert}. Specifically, the text-only QA model trains an extractive answer predictor for the beginning $b$ and ending $e$ of the answer span as follows: $\boldsymbol{\alpha}_b = \text{softmax}(\mathbf{W}_1 \mathbf{H})$ and $\boldsymbol{\alpha}_e = \text{softmax}(\mathbf{W}_2 \mathbf{H})$,
where $\mathbf{W}_1$, $\mathbf{W}_2 \in R^{1\times D}$ and $D$ is the LM's output embedding dimension and $T$ is length of input text. The loss function is the averaged cross entropy on the two answer predictors:
\begin{equation*}
\mathcal{L} = - \frac{1}{2} \sum_{t=1}^{T} \left\{
1(\mathbf{b}_{t}) \log \boldsymbol{\alpha}_{b}^{t} 
+ 1 (\mathbf{e}_t) \log \boldsymbol{\alpha}_{e}^{t} \right\}.
\end{equation*}

\subsection{End-to-End Multimedia QA}

Following recent progress in building pre-trained multimedia models \cite{tan2019lxmert, chen2020uniter, li2020oscar}, we experiment with finetuning an end-to-end multimedia {QA} model for our task. Specifically, we use OSCAR~\cite{li2020oscar} which learns cross-media representations of image-text pairs with object tags from images added as anchor points to significantly ease the learning of alignments. OSCAR has shown SOTA performance in multiple vision-language tasks like VQA~\cite{goyal2017making}, image-text retrieval, image captioning~\cite{you2016image, agrawal2019nocaps} and visual reasoning~\cite{suhr2019corpus}. 

We finetune OSCAR using the synthetic data that we generated in Section \ref{sec:syn_data}. 
We add an extractive answer predictor on top of the outputs from OSCAR to predict the start and end offset of the final answer from the body text. The classifier is trained with cross-entropy loss.

To obtain the image features and the object labels, we use a publicly available implementation\footnote{\url{https://github.com/airsplay/py-bottom-up-attention}} of Faster-RCNN feature extraction tool~\cite{ren2015faster, anderson2018bottom}. 

\subsection{Pipeline-based Multimedia QA}

An overview of our pipeline-based baseline for multi-hop multimedia QA is shown in Figure~\ref{fig:mmqa_pipeline}. First, we split a multi-hop question into a question referencing the image, referred to as \textit{image-question}, and a question about the text, referred to as \textit{text-question}. To achieve this, we use a multi-hop question decomposition model \cite{min2019multi}.
For example, in \figurename~\ref{fig:mmqa_pipeline}, the question ``What did the person with the red coat in the image talk about?'' is decomposed to ``Who is the person with the red coat in the image'' and ``What did [ANSWER] talk about'', where [ANSWER] denotes the answer of the first question. We take the first question as image-question and the second one as text-question. 

Next, we find a bounding box that can answer the image-question, such as the blue bounding box in \figurename~\ref{fig:mmqa_pipeline}. In detail, we first obtain the bounding boxes and identify the visual attributes for these bounding boxes using the approach described in section \ref{sec:visual_attr}.  We then match the image-question to the bounding boxes based on the similarity of their embeddings. The bounding box is represented as bag-of-words over its visual attribute classes, such as ``woman, suit, red''; the question embedding is also represented as bag-of-words over the tokens in the image-question with stop words removed. We average the embeddings from FastText~\cite{mikolov2018advances} and compute cosine similarity. 

Then, we obtain a text span associated with the selected bounding box, such as ``Nikki Haley'' for the blue bounding box. Specifically, we use the cross-modal attentions between the bounding boxes and caption text spans from the grounding approach in section \ref{sec:grounding}. We call this text span as \textit{bridge item}. Finally, we insert the bridge item into the text-question and run it against the single-hop text-only QA model \cite{chakravarti2020towards}, to get the final answer.

\section{Experiments}
\subsection{Settings}

The multi-hop text-only QA model is trained on HotpotQA~\cite{yang2018hotpotqa}, which is a multi-hop extractive QA dataset over Wikipedia passages. For the single-hop text-only QA model, which is used in filtering and the pipeline-based multimedia baseline, we follow \cite{chakravarti2020towards} by training first on SQuAD 2.0 \cite{rajpurkar2018know} and then on Natural Questions \cite{kwiatkowski2019natural}. Both models use bert-large-uncased \cite{devlin2019bert}.

For the end-to-end multimedia QA model, we use  OSCAR-large\footnote{\url{https://github.com/microsoft/Oscar}} as the underlying architecture. We start with the pre-trained model provided by \citet{li2020oscar}. The model is then trained on 20k silver-standard training examples from Section \ref{sec:syn_data}.

\subsection{Results and Analysis}
Table~\ref{tab:exp}  provides the results of various baselines on the dev and test sets of our \taskname{} benchmark. The human baseline performance is evaluated over 70 randomly sampled examples. We use the macro-averaged F$_1$ score of the final answer for evaluation.
The benefit of incorporating multimedia knowledge extraction can be seen from the strong performance of the pipeline-based multimedia QA system.
Interestingly, we see that the end-to-end multimedia QA baseline underperforms the multi-hop text-only system. This could be due to OSCAR being pre-trained with image-caption pairs, making it potentially not suited for reasoning over larger text input (news body text in this case). 

\begin{table}[!htb]
    \centering
    \begin{tabular}{l|c|c}
    \hline
    Model & Dev & Test \\
    \hline
    Multi-hop Text-only QA & 25.6 & 24.8 \\
    End-to-end Multimedia QA & 12.1 & 11.5 \\
    Pipeline-based Multimedia QA & \textbf{37.3} & \textbf{32.6} \\
    \hline
    Human Baseline & - & 66.5 \\
    \hline
    \end{tabular}
    \caption{F1 Performance (\%) of different baselines on the \taskname{} evaluation benchmark.}
    \label{tab:exp}
\end{table}

For the pipeline based multimedia baseline, we analyze performance of the two steps (i.e computing the bridge answer and final answer).  We first compute the F$_1$ score for the intermediate answer, which we call bridge F$_1$. We see that the pipeline-based multimedia QA system has a bridge F$_1$ of 35.3\% and 31.4\% on the dev and test sets respectively, whereas the human baseline has a considerably higher bridge F$_1$ of 78.8\%. This shows that the models are considerably behind humans in identifying the grounded object being referred to in the question, thereby justifying our task. Next, we observe that when the correct bridge item is provided to the text question in the second step (i.e., when bridge F1 = 100\%), the system has a final F1 score of 51.5\% and 49.2\% on the dev and test sets respectively. This suggests that even when the right bridge item is provided, finding the final answer is not straightforward.

\begin{figure}[!htb]
    \centering
    \includegraphics[scale=0.31]{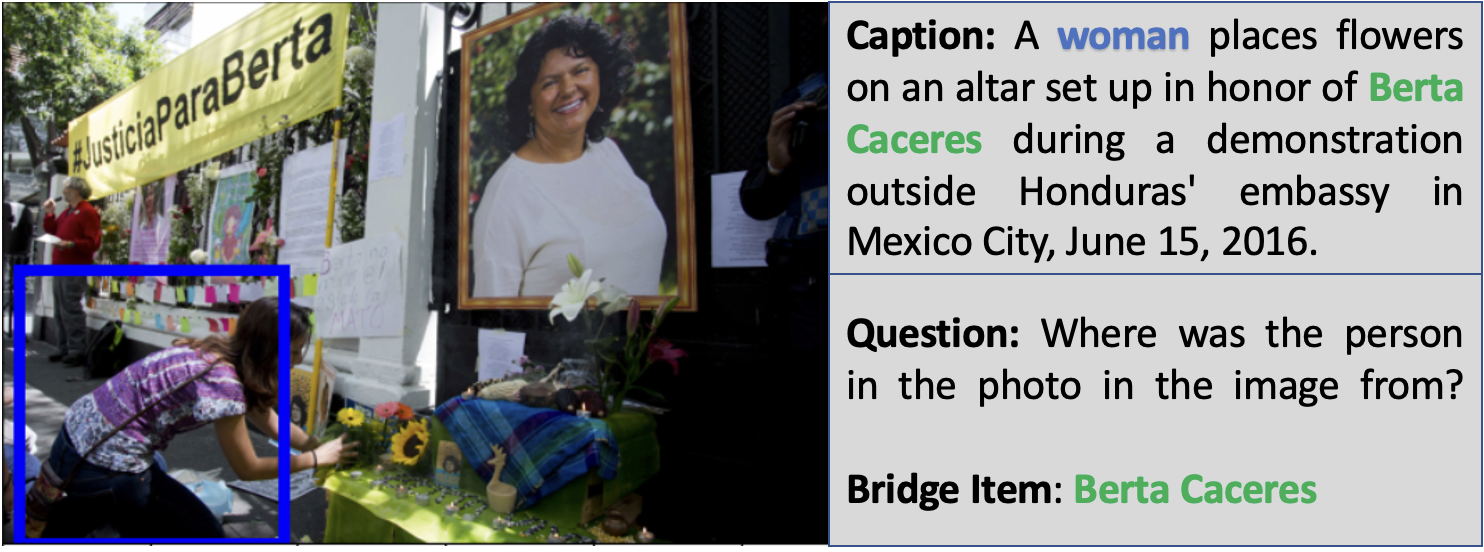}
    \caption{An example where the grounding system failed to capture the ground-truth bridge item (in green). The grounded entity is in blue in the caption and it's corresponding bounding box is shown in blue in the image.}
    \label{fig:grounding_example}
\end{figure}

We analyze performance of individual components of the pipeline based QA baseline, specifically the grounding and attribute matching systems. First, we evaluate the coverage of the bridge item in the output of the grounding system. We see that in 45\% of the cases, the bridge item is present in the canonical mention of the grounded entity. Figure \ref{fig:grounding_example} shows one such example where the grounding system was unable to capture the bridge item. Next, whenever the bridge item is captured in the grounding, we observe that the attribute matching system is able to pick the correct bridge item in 60\% of the cases. 

\subsection{Training Set Analysis}
In this work, we bypass the high cost of manually annotating examples by using a carefully designed pipeline to generate a silver-standard training set for our \taskname{} task. While such automatically generated training data is free and can be scaled to much bigger sizes at no cost, we cannot expect the silver-standard training data to match the quality of human-annotated data. We performed a human evaluation study on 100 examples randomly sampled from the automatically generated training set. We observed that 80\% of questions required using image information, 59\% had the correct bridge answer and had an average score of 2.6 (range of 1-3) for grammaticality. Out of the questions with the correct bridge answer, 64\% had the correct final answer. Upon manual inspection of the incorrect cases, we attributed 46\% of errors to the grounding system, 4\% to the visual-attribute extraction and 33\% to the conditioned-question generator, with the  remaining from incorrectly identifying the image co-reference while editing the questions. From this study, we observe that the automatic training data has some noise, with errors arising from different components of the generation pipeline. While it is not economically feasible to annotate thousands of such examples from scratch, one potential direction is to have annotators correct these automatically generated examples to help improve the quality.

\section{Related Work}

Visual Question Answering~\cite{antol2015vqa} aims to find a natural language answer given a natural language question and an image. Several datasets have been proposed, such as VQA~\cite{agrawal2017vqa, goyal2017making},
DAQUAR~\cite{malinowski2014multi}, COCO-QA~\cite{ren2015exploring}, VCR~\cite{zellers2019recognition}, and PlotQA \cite{methani2020plotqa}, all of which require answering questions about images. However, the questions are multiple-choice or the answers are from a predefined vocabulary with only the information in the image being sufficient to get the correct answer. Simple baseline methods that only use question understanding~\cite{kazemi2017show} or answer option sentiment analysis~\cite{manjunatha2019explicit} have been proven to perform surprisingly well on datasets such as VQA~\cite{antol2015vqa} and VQA2.0~\cite{goyal2017making}. 
But they are unlikely to provide good answers for understanding complex events. Current QA datasets over news involve using just the news body text~\cite{trischler2017newsqa}. In contrast, our benchmark focuses on questions with informative answers that require reasoning across multiple data modalities. 

Recently, there has been some interest in using information from multiple modalities for answering questions. The first step in this direction is ManyModalQA~\cite{hannan2020manymodalqa}, which requires figuring out which modality to use when answering the question. However, all questions are still answerable using a single modality, without any need for cross-media reasoning. MultimodalQA~\cite{talmor2021multimodalqa} extends it by using multi-hop questions that require cross-media reasoning, with each image linked to a Wikipedia entity. In contrast, our dataset requires grounding between the image-caption pairs to identify which objects in the image are being referred to in the question. 
Moreover, the images in \taskname{}, which are from news articles, capture real-world events and are hence more realistic. 

Grounding text mentions to image regions has previously been explored via cross-media attention~\cite{YehNIPS2017,tan2019lxmert, li2020cross, li2020oscar} or learning of an optimal transport plan~\cite{chen2020uniter, chen2020graph}. Different from general text phrases, cross-media entity coreference~\cite{akbari2019multi, li2020gaia} takes text knowledge extraction graph as input and ground the entity with graph context. We are the first to explore cross-media grounding in an extractive QA setting.

\section{Conclusions and Future Work}

We present a new QA task, \taskname {}, along with an evaluation benchmark for multimedia news understanding. The task is challenging in the requirement of cross-media grounding over images, captions, and news body text. We demonstrate the benefit of using multimedia knowledge extraction,  both for generating silver-standard training data and for a pipeline-based multimedia QA system.  
The multimedia baselines are still considerably behind human performance, suggesting ample room for improvement. Future work will incorporate other forms of media in news, such as video and audio, to facilitate information seeking from more comprehensive data sources. Another direction is to infuse the end-to-end multimedia QA system with additional input from the grounding and visual attribute extraction systems.

\section*{Acknowledgements}
We would like to thank Sean Kosman, Rebecca Lee, Kathryn Conger and Martha Palmer for their help on data annotations, and thank Prof. Ernest Davis (NYU) for insightful advice and feedback on our data set and paper. This research is based upon work supported in part by U.S. DARPA AIDA Program No. FA8750-18-2-0014 and U.S. DARPA KAIROS Program No. FA8750-19-2-1004. The views and conclusions contained herein are those of the authors and should not be interpreted as necessarily representing the official policies, either expressed or implied, of DARPA, or the U.S. Government. The U.S. Government is authorized to reproduce and distribute reprints for governmental purposes notwithstanding any copyright annotation therein. 

\bibliography{aaai22}

\begin{thebibliography}{46}
\providecommand{\natexlab}[1]{#1}

\bibitem[{Agrawal et~al.(2017)Agrawal, Lu, Antol, Mitchell, Zitnick, Parikh,
  and Batra}]{agrawal2017vqa}
Agrawal, A.; Lu, J.; Antol, S.; Mitchell, M.; Zitnick, C.~L.; Parikh, D.; and
  Batra, D. 2017.
\newblock Vqa: Visual question answering.
\newblock \emph{International Journal of Computer Vision}, 123(1): 4--31.

\bibitem[{Agrawal et~al.(2019)Agrawal, Desai, Wang, Chen, Jain, Johnson, Batra,
  Parikh, Lee, and Anderson}]{agrawal2019nocaps}
Agrawal, H.; Desai, K.; Wang, Y.; Chen, X.; Jain, R.; Johnson, M.; Batra, D.;
  Parikh, D.; Lee, S.; and Anderson, P. 2019.
\newblock nocaps: novel object captioning at scale.
\newblock In \emph{Proceedings of the IEEE/CVF International Conference on
  Computer Vision}, 8948--8957.

\bibitem[{Akbari et~al.(2019)Akbari, Karaman, Bhargava, Chen, Vondrick, and
  Chang}]{akbari2019multi}
Akbari, H.; Karaman, S.; Bhargava, S.; Chen, B.; Vondrick, C.; and Chang, S.-F.
  2019.
\newblock Multi-level Multimodal Common Semantic Space for Image-Phrase
  Grounding.
\newblock In \emph{Proceedings of the IEEE Conference on Computer Vision and
  Pattern Recognition}, 12476--12486.

\bibitem[{Alberti et~al.(2019)Alberti, Andor, Pitler, Devlin, and
  Collins}]{alberti2019synthetic}
Alberti, C.; Andor, D.; Pitler, E.; Devlin, J.; and Collins, M. 2019.
\newblock Synthetic QA Corpora Generation with Roundtrip Consistency.
\newblock In \emph{Proceedings of the 57th Annual Meeting of the Association
  for Computational Linguistics}, 6168--6173.

\bibitem[{Alberti, Lee, and Collins(2019)}]{alberti2019bert}
Alberti, C.; Lee, K.; and Collins, M. 2019.
\newblock A bert baseline for the natural questions.
\newblock \emph{arXiv preprint arXiv:1901.08634}.

\bibitem[{Anderson et~al.(2018)Anderson, He, Buehler, Teney, Johnson, Gould,
  and Zhang}]{anderson2018bottom}
Anderson, P.; He, X.; Buehler, C.; Teney, D.; Johnson, M.; Gould, S.; and
  Zhang, L. 2018.
\newblock Bottom-up and top-down attention for image captioning and visual
  question answering.
\newblock In \emph{Proceedings of the IEEE Conference on Computer Vision and
  Pattern Recognition}, 6077--6086.

\bibitem[{Antol et~al.(2015)Antol, Agrawal, Lu, Mitchell, Batra,
  Lawrence~Zitnick, and Parikh}]{antol2015vqa}
Antol, S.; Agrawal, A.; Lu, J.; Mitchell, M.; Batra, D.; Lawrence~Zitnick, C.;
  and Parikh, D. 2015.
\newblock VQA: Visual question answering.
\newblock In \emph{Proceedings of the IEEE international conference on computer
  vision}, 2425--2433.

\bibitem[{Buhrmester, Kwang, and Gosling(2016)}]{buhrmester2016amazon}
Buhrmester, M.; Kwang, T.; and Gosling, S.~D. 2016.
\newblock Amazon's Mechanical Turk: A new source of inexpensive, yet
  high-quality data?

\bibitem[{Chakravarti et~al.(2020)Chakravarti, Ferritto, Iyer, Pan, Florian,
  Roukos, and Sil}]{chakravarti2020towards}
Chakravarti, R.; Ferritto, A.; Iyer, B.; Pan, L.; Florian, R.; Roukos, S.; and
  Sil, A. 2020.
\newblock Towards building a Robust Industry-scale Question Answering System.
\newblock In \emph{Proceedings of the 28th International Conference on
  Computational Linguistics: Industry Track}, 90--101.

\bibitem[{Chen et~al.(2020{\natexlab{a}})Chen, Gan, Cheng, Li, Carin, and
  Liu}]{chen2020graph}
Chen, L.; Gan, Z.; Cheng, Y.; Li, L.; Carin, L.; and Liu, J.
  2020{\natexlab{a}}.
\newblock Graph optimal transport for cross-domain alignment.
\newblock In \emph{International Conference on Machine Learning}, 1542--1553.
  PMLR.

\bibitem[{Chen et~al.(2020{\natexlab{b}})Chen, Li, Yu, El~Kholy, Ahmed, Gan,
  Cheng, and Liu}]{chen2020uniter}
Chen, Y.-C.; Li, L.; Yu, L.; El~Kholy, A.; Ahmed, F.; Gan, Z.; Cheng, Y.; and
  Liu, J. 2020{\natexlab{b}}.
\newblock Uniter: Universal image-text representation learning.
\newblock In \emph{European Conference on Computer Vision}, 104--120. Springer.

\bibitem[{Devlin et~al.(2019)Devlin, Chang, Lee, and
  Toutanova}]{devlin2019bert}
Devlin, J.; Chang, M.-W.; Lee, K.; and Toutanova, K. 2019.
\newblock BERT: Pre-training of Deep Bidirectional Transformers for Language
  Understanding.
\newblock In \emph{Proceedings of the 2019 Conference of the North American
  Chapter of the Association for Computational Linguistics: Human Language
  Technologies, Volume 1 (Long and Short Papers)}, 4171--4186.

\bibitem[{Goyal et~al.(2017)Goyal, Khot, Summers-Stay, Batra, and
  Parikh}]{goyal2017making}
Goyal, Y.; Khot, T.; Summers-Stay, D.; Batra, D.; and Parikh, D. 2017.
\newblock Making the V in VQA matter: Elevating the role of image understanding
  in Visual Question Answering.
\newblock In \emph{Proceedings of the IEEE Conference on Computer Vision and
  Pattern Recognition}, 6904--6913.

\bibitem[{Hannan, Jain, and Bansal(2020)}]{hannan2020manymodalqa}
Hannan, D.; Jain, A.; and Bansal, M. 2020.
\newblock ManyModalQA: Modality Disambiguation and QA over Diverse Inputs.
\newblock In \emph{Proceedings of the AAAI Conference on Artificial
  Intelligence}, volume~34, 7879--7886.

\bibitem[{Kazemi and Elqursh(2017)}]{kazemi2017show}
Kazemi, V.; and Elqursh, A. 2017.
\newblock Show, ask, attend, and answer: A strong baseline for visual question
  answering.
\newblock \emph{arXiv preprint arXiv:1704.03162}.

\bibitem[{Kembhavi et~al.(2017)Kembhavi, Seo, Schwenk, Choi, Farhadi, and
  Hajishirzi}]{kembhavi2017you}
Kembhavi, A.; Seo, M.; Schwenk, D.; Choi, J.; Farhadi, A.; and Hajishirzi, H.
  2017.
\newblock Are you smarter than a sixth grader? textbook question answering for
  multimodal machine comprehension.
\newblock In \emph{Proceedings of the IEEE Conference on Computer Vision and
  Pattern recognition}, 4999--5007.

\bibitem[{Kwiatkowski et~al.(2019)Kwiatkowski, Palomaki, Redfield, Collins,
  Parikh, Alberti, Epstein, Polosukhin, Devlin, Lee
  et~al.}]{kwiatkowski2019natural}
Kwiatkowski, T.; Palomaki, J.; Redfield, O.; Collins, M.; Parikh, A.; Alberti,
  C.; Epstein, D.; Polosukhin, I.; Devlin, J.; Lee, K.; et~al. 2019.
\newblock Natural questions: a benchmark for question answering research.
\newblock \emph{Transactions of the Association for Computational Linguistics},
  7: 453--466.

\bibitem[{Lei et~al.(2018)Lei, Yu, Bansal, and Berg}]{lei2018tvqa}
Lei, J.; Yu, L.; Bansal, M.; and Berg, T. 2018.
\newblock TVQA: Localized, Compositional Video Question Answering.
\newblock In \emph{Proceedings of the 2018 Conference on Empirical Methods in
  Natural Language Processing}, 1369--1379.

\bibitem[{Lewis et~al.(2020{\natexlab{a}})Lewis, Liu, Goyal, Ghazvininejad,
  Mohamed, Levy, Stoyanov, and Zettlemoyer}]{lewis2020bart}
Lewis, M.; Liu, Y.; Goyal, N.; Ghazvininejad, M.; Mohamed, A.; Levy, O.;
  Stoyanov, V.; and Zettlemoyer, L. 2020{\natexlab{a}}.
\newblock BART: Denoising Sequence-to-Sequence Pre-training for Natural
  Language Generation, Translation, and Comprehension.
\newblock In \emph{Proceedings of the 58th Annual Meeting of the Association
  for Computational Linguistics}, 7871--7880.

\bibitem[{Lewis et~al.(2020{\natexlab{b}})Lewis, Oguz, Rinott, Riedel, and
  Schwenk}]{lewis2020mlqa}
Lewis, P.; Oguz, B.; Rinott, R.; Riedel, S.; and Schwenk, H.
  2020{\natexlab{b}}.
\newblock MLQA: Evaluating Cross-lingual Extractive Question Answering.
\newblock In \emph{Proceedings of the 58th Annual Meeting of the Association
  for Computational Linguistics}, 7315--7330.

\bibitem[{Li et~al.(2020{\natexlab{a}})Li, Zareian, Lin, Pan, Whitehead, Chen,
  Wu, Ji, Chang, Voss et~al.}]{li2020gaia}
Li, M.; Zareian, A.; Lin, Y.; Pan, X.; Whitehead, S.; Chen, B.; Wu, B.; Ji, H.;
  Chang, S.-F.; Voss, C.; et~al. 2020{\natexlab{a}}.
\newblock Gaia: A fine-grained multimedia knowledge extraction system.
\newblock In \emph{Proceedings of the 58th Annual Meeting of the Association
  for Computational Linguistics: System Demonstrations}, 77--86.

\bibitem[{Li et~al.(2020{\natexlab{b}})Li, Zareian, Zeng, Whitehead, Lu, Ji,
  and Chang}]{li2020cross}
Li, M.; Zareian, A.; Zeng, Q.; Whitehead, S.; Lu, D.; Ji, H.; and Chang, S.-F.
  2020{\natexlab{b}}.
\newblock Cross-media Structured Common Space for Multimedia Event Extraction.
\newblock In \emph{Proceedings of the 58th Annual Meeting of the Association
  for Computational Linguistics}, 2557--2568.

\bibitem[{Li et~al.(2020{\natexlab{c}})Li, Yin, Li, Zhang, Hu, Zhang, Wang, Hu,
  Dong, Wei et~al.}]{li2020oscar}
Li, X.; Yin, X.; Li, C.; Zhang, P.; Hu, X.; Zhang, L.; Wang, L.; Hu, H.; Dong,
  L.; Wei, F.; et~al. 2020{\natexlab{c}}.
\newblock Oscar: Object-semantics aligned pre-training for vision-language
  tasks.
\newblock In \emph{European Conference on Computer Vision}, 121--137. Springer.

\bibitem[{Lin et~al.(2019)Lin, Zheng, Zheng, Wu, Hu, Yan, and
  Yang}]{lin2019improving}
Lin, Y.; Zheng, L.; Zheng, Z.; Wu, Y.; Hu, Z.; Yan, C.; and Yang, Y. 2019.
\newblock Improving person re-identification by attribute and identity
  learning.
\newblock \emph{Pattern Recognition}, 95: 151--161.

\bibitem[{Malinowski and Fritz(2014)}]{malinowski2014multi}
Malinowski, M.; and Fritz, M. 2014.
\newblock A multi-world approach to question answering about real-world scenes
  based on uncertain input.
\newblock In \emph{Advances in neural information processing systems},
  1682--1690.

\bibitem[{Manjunatha, Saini, and Davis(2019)}]{manjunatha2019explicit}
Manjunatha, V.; Saini, N.; and Davis, L.~S. 2019.
\newblock Explicit Bias Discovery in Visual Question Answering Models.
\newblock In \emph{Proceedings of the IEEE Conference on Computer Vision and
  Pattern Recognition}, 9562--9571.

\bibitem[{Methani et~al.(2020)Methani, Ganguly, Khapra, and
  Kumar}]{methani2020plotqa}
Methani, N.; Ganguly, P.; Khapra, M.~M.; and Kumar, P. 2020.
\newblock Plotqa: Reasoning over scientific plots.
\newblock In \emph{Proceedings of the IEEE/CVF Winter Conference on
  Applications of Computer Vision}, 1527--1536.

\bibitem[{Mikolov et~al.(2018)Mikolov, Grave, Bojanowski, Puhrsch, and
  Joulin}]{mikolov2018advances}
Mikolov, T.; Grave, {\'E}.; Bojanowski, P.; Puhrsch, C.; and Joulin, A. 2018.
\newblock Advances in Pre-Training Distributed Word Representations.
\newblock In \emph{Proceedings of the Eleventh International Conference on
  Language Resources and Evaluation (LREC 2018)}.

\bibitem[{Min et~al.(2019)Min, Zhong, Zettlemoyer, and
  Hajishirzi}]{min2019multi}
Min, S.; Zhong, V.; Zettlemoyer, L.; and Hajishirzi, H. 2019.
\newblock Multi-hop Reading Comprehension through Question Decomposition and
  Rescoring.
\newblock In \emph{Proceedings of the 57th Annual Meeting of the Association
  for Computational Linguistics}, 6097--6109.

\bibitem[{Peters et~al.(2018)Peters, Neumann, Iyyer, Gardner, Clark, Lee, and
  Zettlemoyer}]{peters2018deep}
Peters, M.; Neumann, M.; Iyyer, M.; Gardner, M.; Clark, C.; Lee, K.; and
  Zettlemoyer, L. 2018.
\newblock Deep Contextualized Word Representations.
\newblock In \emph{Proceedings of the 2018 Conference of the North American
  Chapter of the Association for Computational Linguistics: Human Language
  Technologies, Volume 1 (Long Papers)}, 2227--2237.

\bibitem[{Rajpurkar, Jia, and Liang(2018)}]{rajpurkar2018know}
Rajpurkar, P.; Jia, R.; and Liang, P. 2018.
\newblock Know What You Don’t Know: Unanswerable Questions for SQuAD.
\newblock In \emph{Proceedings of the 56th Annual Meeting of the Association
  for Computational Linguistics (Volume 2: Short Papers)}, 784--789.

\bibitem[{Rajpurkar et~al.(2016)Rajpurkar, Zhang, Lopyrev, and
  Liang}]{rajpurkar2016squad}
Rajpurkar, P.; Zhang, J.; Lopyrev, K.; and Liang, P. 2016.
\newblock SQuAD: 100,000+ Questions for Machine Comprehension of Text.
\newblock In \emph{Proceedings of the 2016 Conference on Empirical Methods in
  Natural Language Processing}, 2383--2392.

\bibitem[{Ren, Kiros, and Zemel(2015)}]{ren2015exploring}
Ren, M.; Kiros, R.; and Zemel, R.~S. 2015.
\newblock Exploring models and data for image question answering.
\newblock In \emph{Proceedings of the 28th International Conference on Neural
  Information Processing Systems-Volume 2}, 2953--2961.

\bibitem[{Ren et~al.(2015)Ren, He, Girshick, and Sun}]{ren2015faster}
Ren, S.; He, K.; Girshick, R.; and Sun, J. 2015.
\newblock Faster r-cnn: Towards real-time object detection with region proposal
  networks.
\newblock In \emph{Advances in neural information processing systems}, 91--99.

\bibitem[{Shakeri et~al.(2020)Shakeri, dos Santos, Zhu, Ng, Nan, Wang,
  Nallapati, and Xiang}]{shakeri2020end}
Shakeri, S.; dos Santos, C.; Zhu, H.; Ng, P.; Nan, F.; Wang, Z.; Nallapati, R.;
  and Xiang, B. 2020.
\newblock End-to-End Synthetic Data Generation for Domain Adaptation of
  Question Answering Systems.
\newblock In \emph{Proceedings of the 2020 Conference on Empirical Methods in
  Natural Language Processing (EMNLP)}, 5445--5460.

\bibitem[{Suhr et~al.(2019)Suhr, Zhou, Zhang, Zhang, Bai, and
  Artzi}]{suhr2019corpus}
Suhr, A.; Zhou, S.; Zhang, A.; Zhang, I.; Bai, H.; and Artzi, Y. 2019.
\newblock A Corpus for Reasoning about Natural Language Grounded in
  Photographs.
\newblock In \emph{Proceedings of the 57th Annual Meeting of the Association
  for Computational Linguistics}, 6418--6428.

\bibitem[{Talmor et~al.(2021)Talmor, Yoran, Catav, Lahav, Wang, Asai, Ilharco,
  Hajishirzi, and Berant}]{talmor2021multimodalqa}
Talmor, A.; Yoran, O.; Catav, A.; Lahav, D.; Wang, Y.; Asai, A.; Ilharco, G.;
  Hajishirzi, H.; and Berant, J. 2021.
\newblock MultiModalQA: Complex Question Answering over Text, Tables and
  Images.
\newblock \emph{arXiv preprint arXiv:2104.06039}.

\bibitem[{Tan and Bansal(2019)}]{tan2019lxmert}
Tan, H.; and Bansal, M. 2019.
\newblock LXMERT: Learning Cross-Modality Encoder Representations from
  Transformers.
\newblock In \emph{Proceedings of the 2019 Conference on Empirical Methods in
  Natural Language Processing and the 9th International Joint Conference on
  Natural Language Processing (EMNLP-IJCNLP)}, 5103--5114.

\bibitem[{Trischler et~al.(2017)Trischler, Wang, Yuan, Harris, Sordoni,
  Bachman, and Suleman}]{trischler2017newsqa}
Trischler, A.; Wang, T.; Yuan, X.; Harris, J.; Sordoni, A.; Bachman, P.; and
  Suleman, K. 2017.
\newblock NewsQA: A Machine Comprehension Dataset.
\newblock In \emph{Proceedings of the 2nd Workshop on Representation Learning
  for NLP}, 191--200.

\bibitem[{Welbl, Stenetorp, and Riedel(2018)}]{welbl2018constructing}
Welbl, J.; Stenetorp, P.; and Riedel, S. 2018.
\newblock Constructing datasets for multi-hop reading comprehension across
  documents.
\newblock \emph{Transactions of the Association for Computational Linguistics},
  6: 287--302.

\bibitem[{Yang et~al.(2003)Yang, Chaisorn, Zhao, Neo, and
  Chua}]{yang2003videoqa}
Yang, H.; Chaisorn, L.; Zhao, Y.; Neo, S.-Y.; and Chua, T.-S. 2003.
\newblock VideoQA: question answering on news video.
\newblock In \emph{Proceedings of the eleventh ACM international conference on
  Multimedia}, 632--641. ACM.

\bibitem[{Yang et~al.(2018)Yang, Qi, Zhang, Bengio, Cohen, Salakhutdinov, and
  Manning}]{yang2018hotpotqa}
Yang, Z.; Qi, P.; Zhang, S.; Bengio, Y.; Cohen, W.; Salakhutdinov, R.; and
  Manning, C.~D. 2018.
\newblock HotpotQA: A Dataset for Diverse, Explainable Multi-hop Question
  Answering.
\newblock In \emph{Proceedings of the 2018 Conference on Empirical Methods in
  Natural Language Processing}, 2369--2380.

\bibitem[{Yeh et~al.(2017)Yeh, Xiong, mei Hwu, Do, and Schwing}]{YehNIPS2017}
Yeh, R.; Xiong, J.; mei Hwu, W.; Do, M.; and Schwing, A. 2017.
\newblock {Interpretable and Globally Optimal Prediction for Textual Grounding
  using Image Concepts}.
\newblock In \emph{Proc. Neural Information Processing Systems}.

\bibitem[{You et~al.(2016)You, Jin, Wang, Fang, and Luo}]{you2016image}
You, Q.; Jin, H.; Wang, Z.; Fang, C.; and Luo, J. 2016.
\newblock Image captioning with semantic attention.
\newblock In \emph{Proceedings of the IEEE conference on computer vision and
  pattern recognition}, 4651--4659.

\bibitem[{Zellers et~al.(2019)Zellers, Bisk, Farhadi, and
  Choi}]{zellers2019recognition}
Zellers, R.; Bisk, Y.; Farhadi, A.; and Choi, Y. 2019.
\newblock From recognition to cognition: Visual commonsense reasoning.
\newblock In \emph{Proceedings of the IEEE/CVF Conference on Computer Vision
  and Pattern Recognition}, 6720--6731.

\bibitem[{Zheng et~al.(2015)Zheng, Shen, Tian, Wang, Wang, and
  Tian}]{zheng2015scalable}
Zheng, L.; Shen, L.; Tian, L.; Wang, S.; Wang, J.; and Tian, Q. 2015.
\newblock Scalable person re-identification: A benchmark.
\newblock In \emph{Proceedings of the IEEE international conference on computer
  vision}, 1116--1124.

\end{thebibliography}

\appendix

\section{Appendix}

\subsection{Annotation interface}

\begin{figure*}
    \centering
    \includegraphics[width=0.77\textwidth]{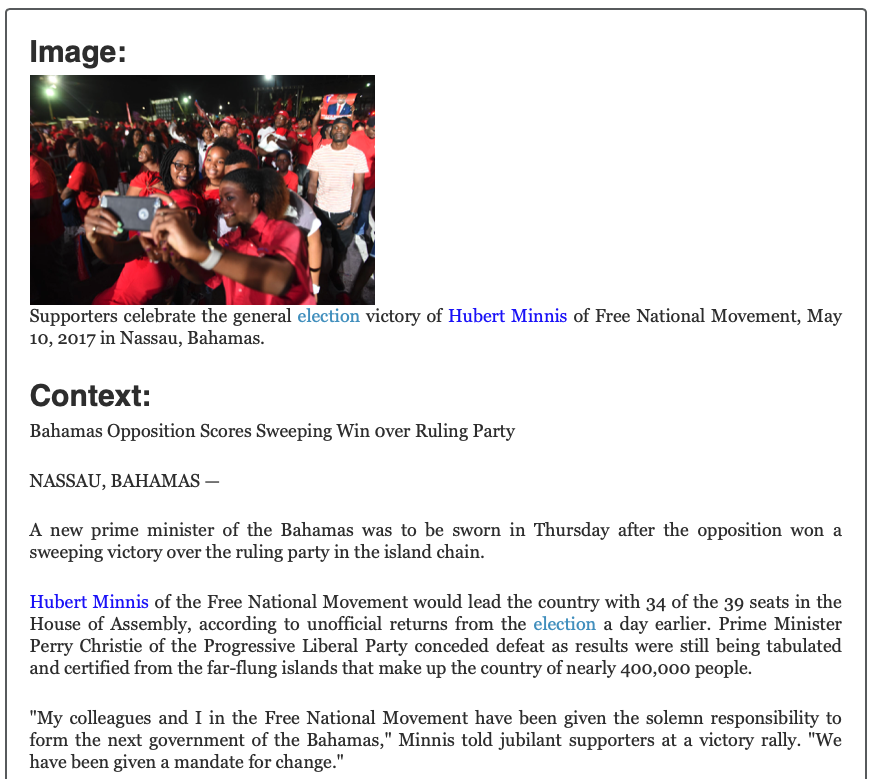}
    \includegraphics[width=0.77\textwidth]{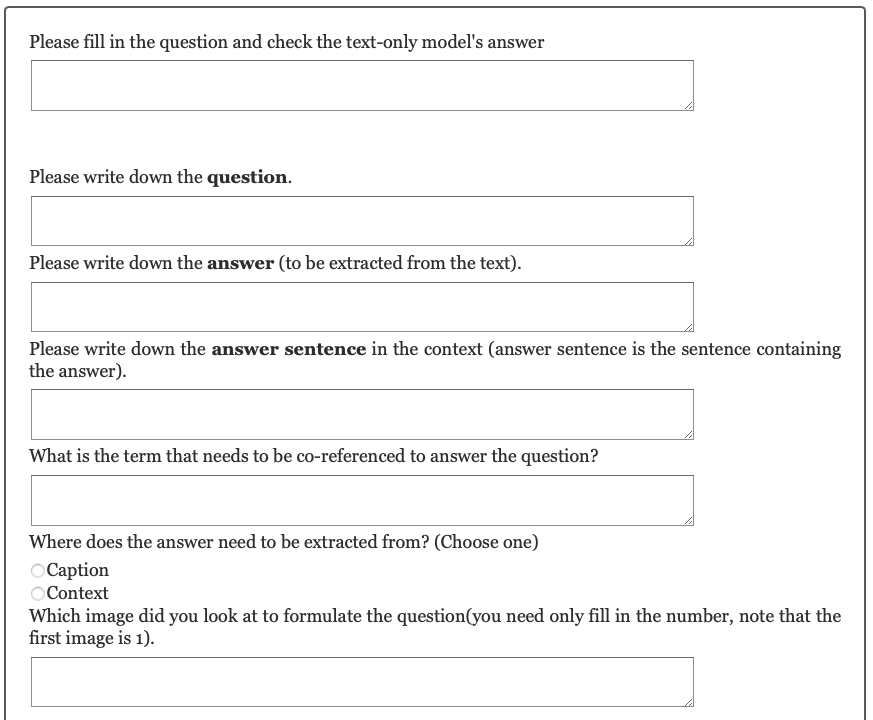}

    \caption{Screenshot of the annotation interface.}
    \label{fig:annotation_interface}
\end{figure*}

We collected our \taskname{} evaluation set using a specially tailored annotation interface, which shows news articles along with image-caption pairs. For automatic quality control, the interface also provides an access to a single-hop text-only QA model so that we can prevent annotators from writing questions that can be directly answered using news body text. Screenshot of the annotation interface is shown in Figure~\ref{fig:annotation_interface}.

\subsubsection{Instructions}

We provide annotators with the following list of steps to follow while coming up with the questions:

\begin{enumerate}
    \item Look at the image-caption pairs and briefly read through the article to get a sense of what the news article is about.
    
   \item Pick an \textit{informative} image you want to ask the question about. Informative image is one which has some events happening that are about the topic of the news article. 

\item After deciding on the image-caption pair, look for an entity in the caption that is grounded in the image. This is the co-referential item which the question will be about.

\item Next, search for appearances of this co-referential object in the news article text and pick a sentence from which you want to form a question. 

\item Now formulate a question using this sentence. Refer to the co-referential item with respect to it’s connection with the image.

\item Check whether the question can be answered by a text-only model. If the top-1 answer from the text-only model get’s part of your answer right or if it comes from your answer sentence, then consider that a text-only model is able to answer your question.

\item If a text-only model is able to answer your question, try changing the question or try picking a different answer sentence to ask a question about. You can also pick a different co-referential item

\end{enumerate}

\subsection{Visual Attribute Extraction}
\label{sec:visual_attr}

\begin{figure}[!htb]
\centering
  \includegraphics[width=1.0\linewidth]{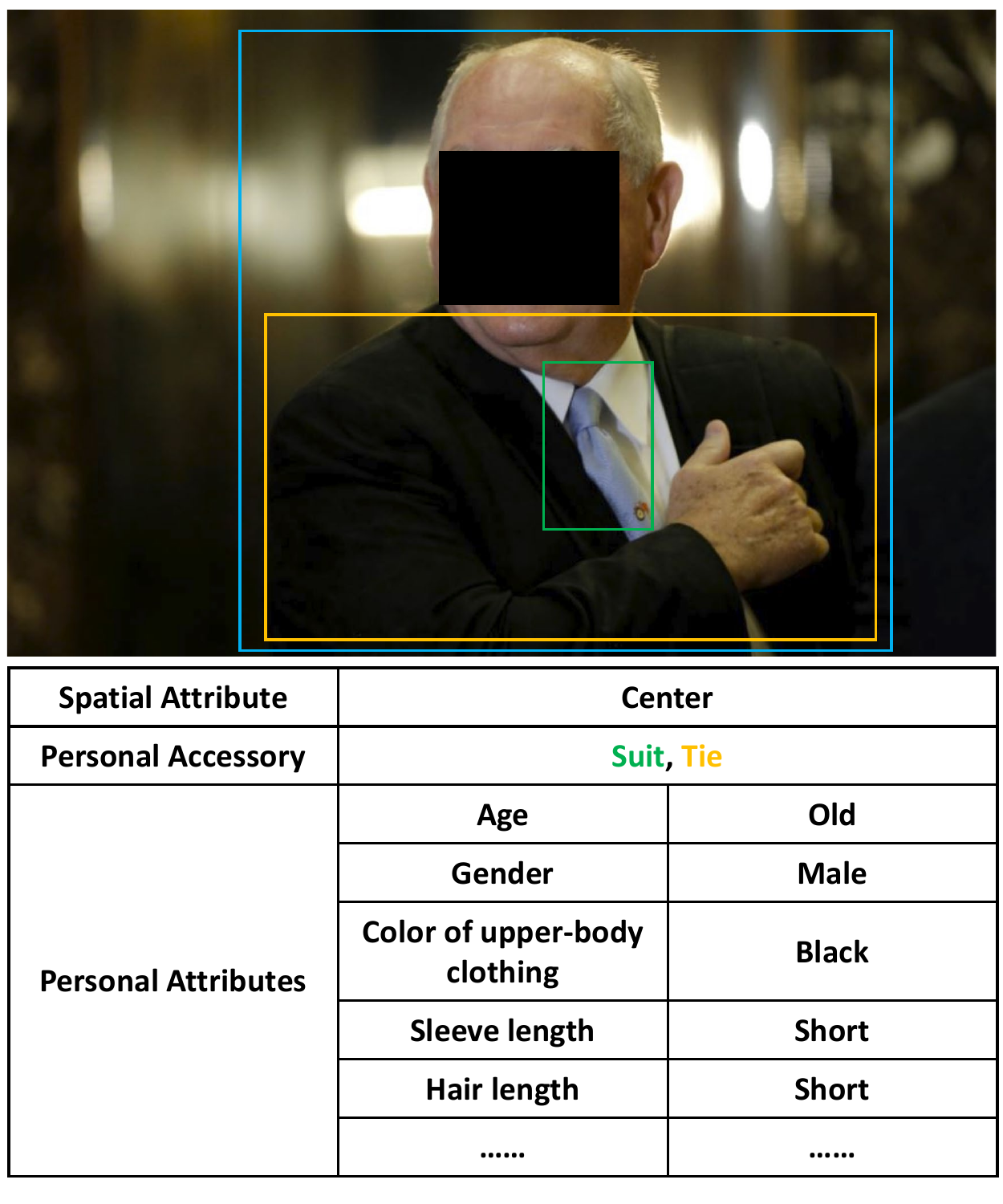}

\caption{Illustration of visual attributes we extract. We mask the face (black box) for privacy.
}
\label{fig:visual_Attribute}
\end{figure}

Visual attributes are extracted as shown in Figure \ref{fig:visual_Attribute}. Detailed extraction strategies are as follows.

The spatial attribute is the relative position $\{$\textit{right, left, top, bottom, center}$\}$ of the person in the image, which is determined by the center $(x,y)$ of the bounding box, compared to the center of the image. If $x > 0.7*w$, then right is returned; if $x < 0.3*w$, then left is returned. If $y > 0.7*h$, then bottom is returned; if $y < 0.3*h$, then top is returned. If none of above is returned, then return center.

The association process of personal accessories is rule-based: if $\frac{Area_{P\bigcap A}}{Area_A}>0.9$, then we assign the accessory to the person. Here, $P$ and $A$ represent the person bounding box and accessory bounding box respectively. We identify the following 15 personal accessories: coat, tie, jacket, suit, dress, sunglasses, glasses, hat, shirt, helmet, goggles, handbag, scarf, glove, life jacket.

\vspace{4em}

\begin{table}[t]
    \centering
    \begin{tabular}{|c|p{9em}|}
    \hline
          \textbf{Category} & \textbf{Possible Labels} \\\hline
          Gender & Male, Female \\\hline
          Age & Young, Teenager, Adult, Old \\\hline
          Hair length & Short, Long \\\hline
          Sleeve length & Short, Long \\\hline
          Length of lower-body clothing & Short, Long \\\hline
          Type of lower-body clothing & Dress, Pants \\\hline
          Wearing hat & No, Yes \\\hline
          Carrying backpack & No, Yes \\\hline
          Carrying bag & No, Yes \\\hline
          Carrying handbag & No, Yes \\\hline
         Color of upper-body clothing & Black, White, Red, Purple, Yellow, Gray, Blue, Green \\\hline
         Color of lower-body clothing & Black, White, Pink, Purple, Yellow, Gray, Blue, Green, Brown \\\hline
    \end{tabular}
    \caption{Person Attributes on the Market-1501 dataset.}
    \label{tab:attri}
\end{table}

The personal attribute recognition model \cite{lin2019improving} is trained on a benchmark dataset, Market-1501 \cite{zheng2015scalable}, achieving 94\% test accuracy. The dataset has 27 attribute types distributed in 12 categories, as shown in Table~\ref{tab:attri}. 

\subsection{Silver-standard Training Examples}

\begin{figure}[!htb]
    \centering
    \includegraphics[scale=0.31]{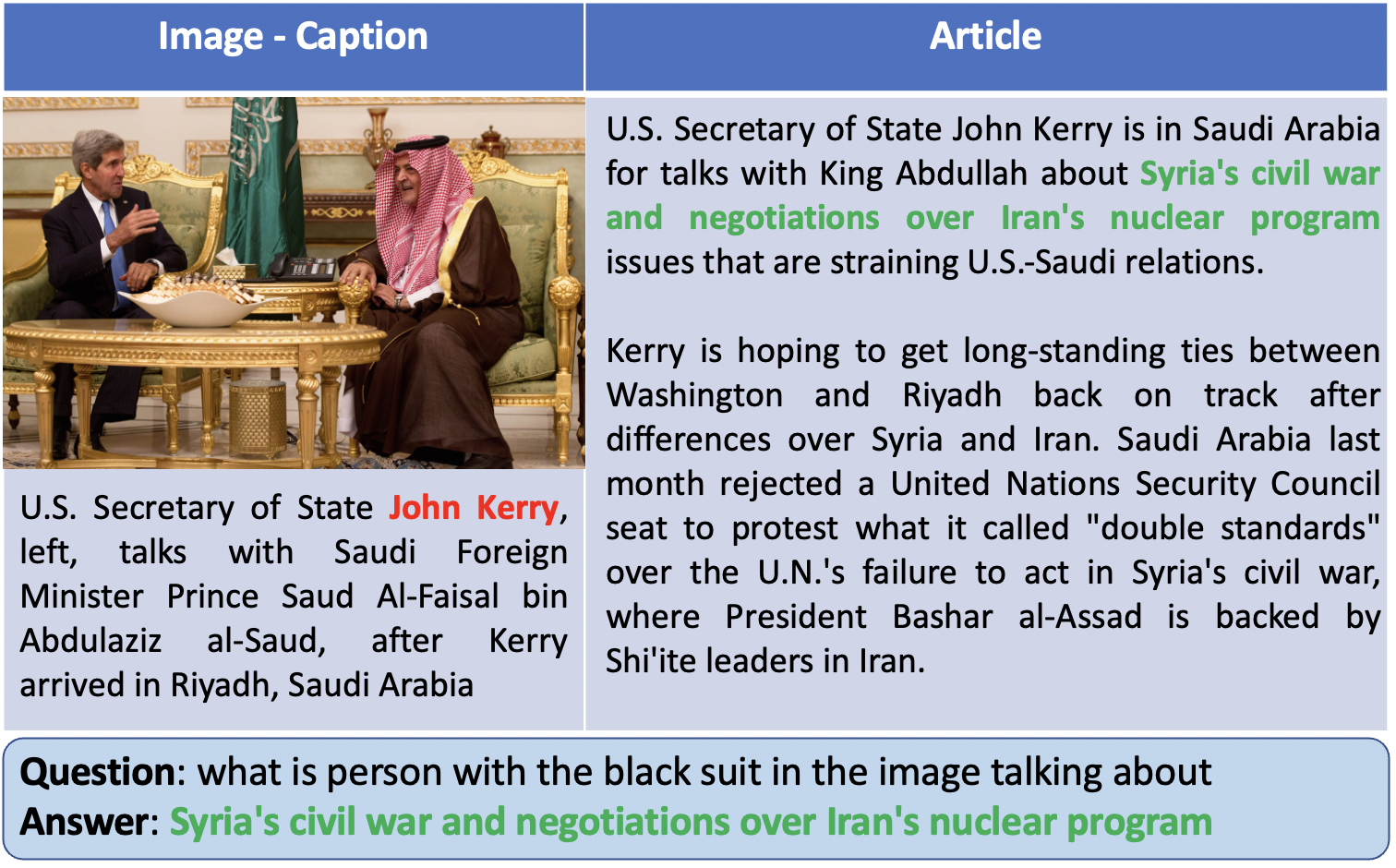}
    \caption{An example from the silver-standard training set. The bridge item is shown in the caption in red and the answer is shown in green in the body text}
    \label{fig:ex_1}
\end{figure}

\begin{figure}[!htb]
    \centering
    \includegraphics[scale=0.31]{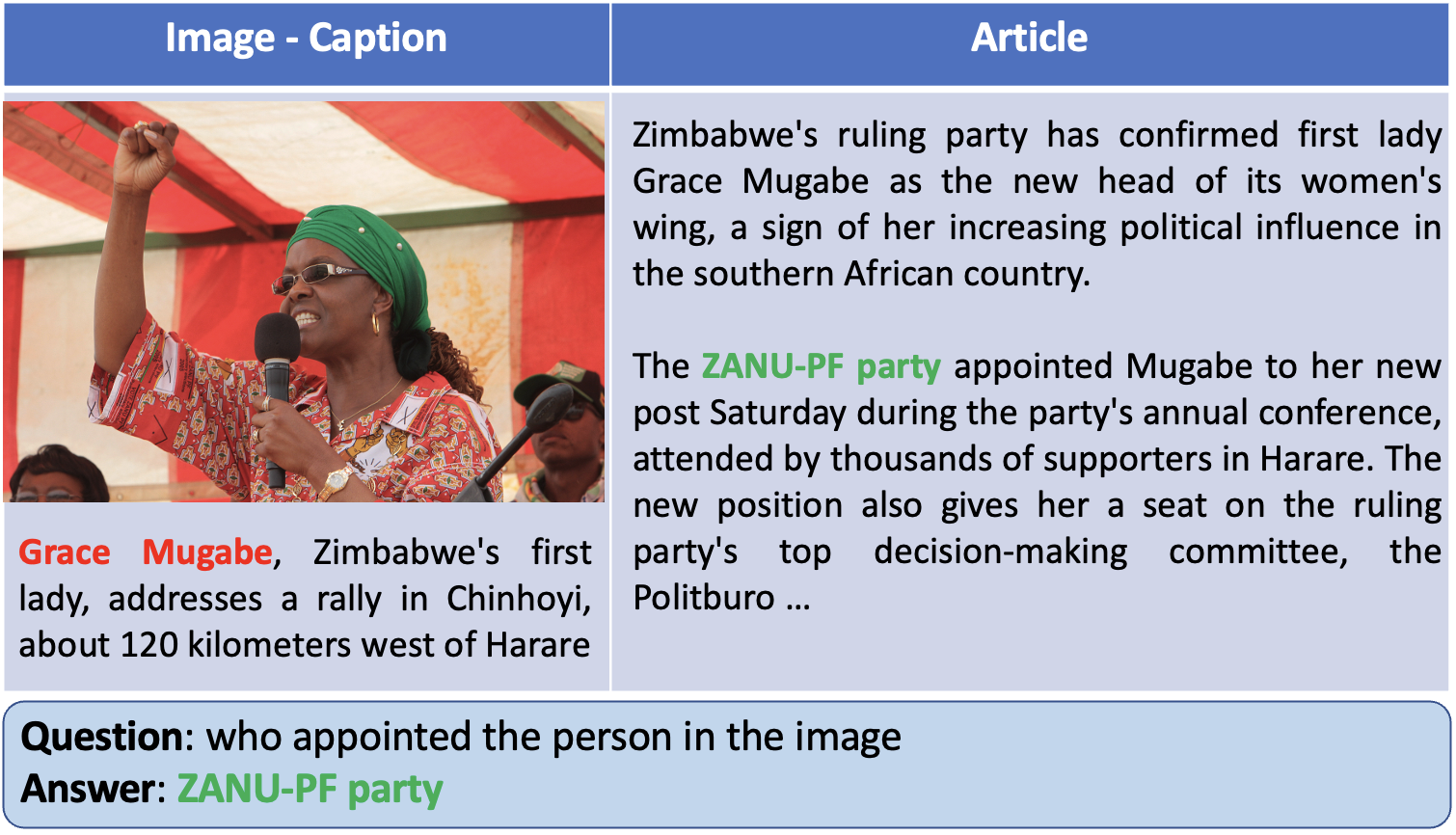}
    \caption{An example from the silver-standard training set. The bridge item is shown in the caption in red and the answer is shown in green in the body text}
    \label{fig:ex_2}
\end{figure}

Figures \ref{fig:ex_1}, \ref{fig:ex_2} show some silver-standard training examples that are output from the synthetic generation pipeline. 

\end{document}